\documentclass{article}
\usepackage{ijcai25}

\usepackage{times}
\usepackage{soul}
\usepackage{url}
\usepackage[hidelinks]{hyperref}
\usepackage[utf8]{inputenc}
\usepackage[small]{caption}
\usepackage{graphicx}
\usepackage{amsmath}
\usepackage{amsthm}
\usepackage{booktabs}
\usepackage{algorithm}
\usepackage{algorithmic}
\usepackage[switch]{lineno}

\usepackage{multirow}
\usepackage{colortbl}
\usepackage[table]{xcolor}
\usepackage{amsfonts}
\title{Efficient Visual Representation Learning with Heat Conduction Equation}

\author{
Zhemin Zhang$^1$
\and
Xun Gong$^{1}$ \\
\affiliations
$^1$School of Computing and Artificial Intelligence, Southwest Jiaotong University\\
\emails
 zheminzhang@my.swjtu.edu.cn, 
 xgong@swjtu.edu.cn
}

\begin{document}

\maketitle

\begin{abstract}
Foundation models, such as CNNs and ViTs, have powered the development of image representation learning. However, general guidance to model architecture design is still missing. Inspired by the connection between image representation learning and heat conduction, we model images by the heat conduction equation, where the essential idea is to conceptualize image features as temperatures and model their information interaction as the diffusion of thermal energy. Based on this idea, we find that many modern model architectures, such as residual structures, SE block, and feed-forward networks, can be interpreted from the perspective of the heat conduction equation. Therefore, we leverage the heat equation to design new and more interpretable models. As an example, we propose the Heat Conduction Layer and the Refinement Approximation Layer inspired by solving the heat conduction equation using Finite Difference Method and Fourier series, respectively. The main goal of this paper is to integrate the overall architectural design of neural networks into the theoretical framework of heat conduction. Nevertheless, our Heat Conduction Network (HcNet) still shows competitive performance, e.g., HcNet-T achieves 83.0\% top-1 accuracy on ImageNet-1K while only requiring 28M parameters and 4.1G MACs. The code is publicly available at: https://github.com/ZheminZhang1/HcNet.
\end{abstract}

\section{Introduction}

The dominance of convolutional neural networks (CNNs) in computer vision has been long-standing. Transformers \cite{NIPS2017-3f5ee243}, originated from natural language processing (NLP), have recently demonstrated state-of-the-art performance in visual learning. Despite tremendous successes, the full self-attention has computational and memory requirement that is quadratic in the sequence length, thereby limiting its applicability to high-resolution vision tasks. Moreover, modeling 2-D images as sequences impairs interpretability. Recently, using classical physical models such as state space \cite{gu2024mambalineartimesequencemodeling,liu2024vmambavisualstatespace}, gravitational field \cite{NEURIPS2022-6ad68a54}, and heat conduction equation \cite{chen2022selfsupervisedlearningbasedheat} to inspire architecture design has become an effective way to explore new model architectures. 

\begin{figure}[t]
\centering
\includegraphics[width=1.0\linewidth]{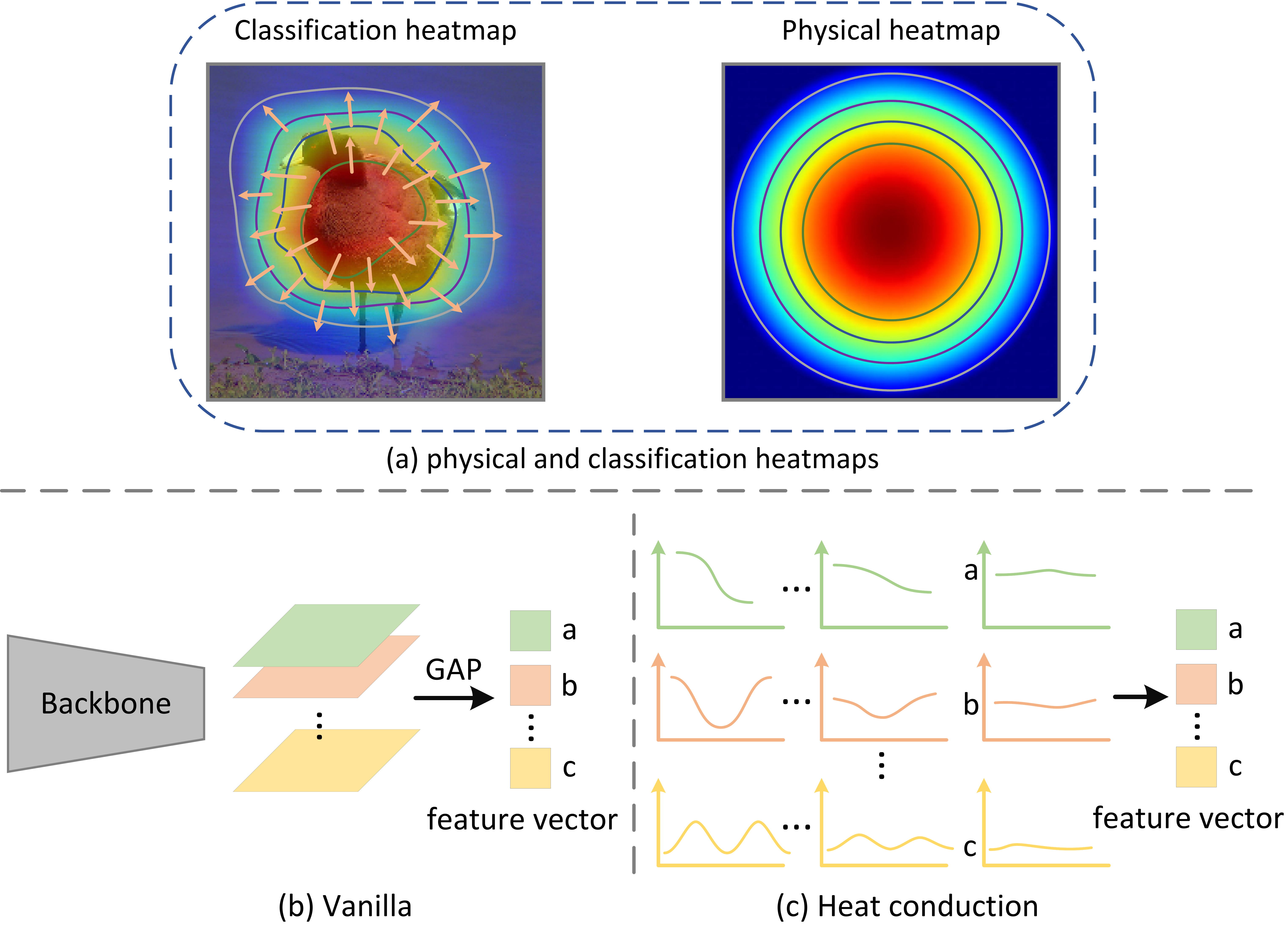} % Reduce the figure size so that it is slightly narrower than the column.
\caption{The connection between vanilla image representation learning and heat conduction. \textbf{(a)} Comparison of physical and classification heatmaps. \textbf{(b)} Vanilla feature extraction process. GAP denotes global average pooling and a, b, and c are scalars. \textbf{(c)} Thermal diffusion process. The temperature distribution gradually approaches a steady state over time. This process is similar to information compression and abstraction in vanilla feature extraction. Note that, for the sake of clarity, we use 1D temperature distributions as examples, and 2D temperature distributions are similar.}
\label{PhysicalHeatmap-flabel}
\end{figure}

Our motivation is primarily driven by two key observations. First, the physical heatmap computed from heat equation and the classification heatmap learned from supervised learning are very similar, as shown in Figure \ref{PhysicalHeatmap-flabel}(a). This may be due to the propagation of thermal energy through the collision of neighboring particles during heat conduction, resulting in similar temperatures in local space, which is similar to the property that neighboring regions tend to contain related information in images. Secondly, the vanilla feature extraction process maps the image to a low-dimensional feature space, which can be regarded as the compression and abstraction of the image. In this process, the model retains the most important and discriminative global information in the image, while ignoring redundant details, as shown in Figure \ref{PhysicalHeatmap-flabel}(b). This mechanism is similar to heat conduction, where high-frequency information gradually decays, and low-frequency global information becomes dominant over time, as shown in Figure \ref{PhysicalHeatmap-flabel}(c). Based on the above inspiration, connecting heat conduction and visual semantic propagation, we propose HcNet. This work shows that many modern model components, such as residual structures \cite{He-2016-CVPR}, SE block \cite{Hu-2018-CVPR} and feed-forward network (FFN) \cite{NIPS2017-3f5ee243}, can be interpreted in terms of the heat conduction equation. From the perspective of the heat equation, the success of deep neural networks is mainly due to their ability to efficiently approximate heat conduction processes with complex initial conditions.

By establishing a connection between model architecture and the heat conduction equation, we can take advantage of the rich knowledge in the heat conduction equation to guide us in designing novel and more interpretable models. As an example, we propose the Heat Conduction Layer and the Refinement Approximation Layer inspired by solving the heat conduction equation using Finite Difference Method and Fourier series, respectively. Both layers can be explained mathematically using the concept of discrete heat conduction equation. Building upon the aforementioned components, we design a general vision backbone with a hierarchical architecture, named as Heat Conduction Network (HcNet). Compared to vision backbones with various architectures, HcNet consistently achieves competitive performance on image classification, object detection, and semantic segmentation across model scales. Specifically, HcNet-Base achieves 84.3\% top-1 accuracy on ImageNet-1K, surpassing Swin \cite{Liu-2021-ICCV} by 0.8\%. In addition, HcNet has lower computational cost compared to ViT-based models.

Our contributions can be summarized as follows:

\begin{itemize}
    \item We propose the Heat Conduction Layer, a network layer inspired by the finite difference solution of the heat conduction equation, which follows the heat conduction principle to transfer information with low computational complexity and high interpretability.
    \item We propose the Refinement Approximation Layer, a network layer inspired by the Fourier series solution of the heat conduction equation, which improves the fitting ability of the model by linearly combining the features of different channels. The Refinement Approximation Layer explains the role of the FFN layer in terms of Fourier series approximation.
    \item Unlike previous work that applied the heat conduction equation to a specific module of the model, HcNet attempts to guide the design of all model components through the heat conduction equation. This approach helps to introduce rich knowledge in partial differential equations into neural networks.
\end{itemize}

\section{Related Work}

\paragraph{Convolution Neural Networks (CNNs).} In the early eras, ConvNet \cite{lecun1998gradient} serves as the de-facto standard network design for computer vision. The distinctive characteristics of CNNs are encapsulated in the convolution kernels, which enjoy high computational efficiency given specifically designed GPUs. Therefore, many convolutional neural architectures \cite{Radosavovic-2020-CVPR,pmlr-v97-tan19a,Radosavovic-2020-CVPR,pmlr-v139-tan21a} have been proposed as the vision backbone for various visual applications. To address the born limitations of local receptive fields, recent works \cite{Chen-2024-CVPR,Ding-2024-CVPR} have been focused on developing large convolution kernels to obtain larger receptive fields. 

\paragraph{Vision Transformers (ViTs).} Transformers were proposed by Vaswani et al. \cite{NIPS2017-3f5ee243} for machine translation, and have since become the state-of-the-art method in many NLP tasks. The pioneering work, Vision Transformer \cite{dosovitskiy2021imageworth16x16words} demonstrates that pure Transformer-based architectures can also achieve very competitive results in computer vision. One challenge for Vision Transformer-based models is data efficiency. Although ViT can perform better than convolutional networks with hundreds of millions of images for pre-training, such a data requirement is difficult to meet in many cases. To improve data efficiency, many recent works \cite{Liu-2021-ICCV,Tian-2023-CVPR,Zhu-2023-CVPR,10366193,10384565} have focused on introducing the locality and hierarchical structure into ViT. On the other hand, many works have blended attention model with convolution, and accordingly propose a series of hybrid Vision Transformers \cite{hatamizadeh2023fastervit,yu2023metaformer}. 

\paragraph{Physics Inspired Models (PIMs).} The principles of physics bring a novel perspective to foundation model research, taking advantage of the rich knowledge in physics to guide us in designing new and potentially more effective models. The diffusion model \cite{NEURIPS2022-6ad68a54}, motivated by the gravitational field, interprets the data points as electrical charges on a hyperplane and devises a backward ODE to generate the image. Mamba \cite{gu2024mambalineartimesequencemodeling} models sequences based on state space, and Vision Mamba \cite{pmlr-v235-zhu24f,liu2024vmambavisualstatespace} introduces state space into image modeling inspired by mamba. QB-Heat \cite{chen2022selfsupervisedlearningbasedheat} extends the heat equation into high dimensional feature space to predict surrounding image patches for masked image modeling self-supervised task.

Unlike previous work that applied the heat conduction equation to a specific module of the model, HcNet attempts to guide the design of all model components through the heat conduction equation.

\section{Methodology}

\subsection{Preliminaries: Heat Conduction Equation}

Compared with the simple homogeneous diffusion process, the anisotropic heat equation can control the diffusion intensity in different directions, thereby enabling a more precise simulation of the complex heat conduction process. This capability establishes a foundation for the model to effectively capture complex semantic information within images. Considering that images are typically anisotropic, we employ the anisotropic heat equation \cite{widder1976heat} for modeling:
\begin{equation}
\frac{\partial u}{\partial t}={{\alpha }_{x}}\frac{{{\partial }^{2}}u}{\partial {{x}^{2}}}+{{\alpha }_{y}}\frac{{{\partial }^{2}}u}{\partial {{y}^{2}}}
\label{HeatEquation}
\end{equation}
where $u\left( x,y,t \right)$ is the temperature of point $\left( x,y \right)$ at time $t$ within a 2-D region, ${{\alpha }_{x}}$ and ${{\alpha }_{y}}$ denote the thermal diffusivity \cite{bird2002transport} of the material in the $x$- and $y$-directions, respectively.

\paragraph{Discretization.} We use the Finite Difference Method (FDM) to solve the Heat Conduction Equation approximately. Divide the spatial domain into a grid of points. Let the grid points be $\left( {{x}_{i}},{{y}_{j}} \right)$, where $i=1,2,\cdots ,{{N}_{x}}$ and $j=1,2,\cdots ,{{N}_{y}}$. The time domain is also discretized into time steps ${{t}^{n}}$, where $n=0,1,2,\cdots ,{{N}_{t}}$. Substitute the finite difference approximations into the heat conduction equation:
\begin{equation}
\resizebox{1.0\linewidth}{!}{%
$\displaystyle
\begin{split}
\frac{u_{i,j}^{n+1}-u_{i,j}^{n}}{\Delta t} &= \frac{{{\alpha }_{x,1}}(u_{i+1,j}^{n}-u_{i,j}^{n})+{{\alpha }_{x,2}}(u_{i-1,j}^{n}-u_{i,j}^{n})}{\Delta {{x}^{2}}} \\
&\quad + \frac{{{\alpha }_{y,1}}(u_{i,j+1}^{n}-u_{i,j}^{n})+{{\alpha }_{y,2}}(u_{i,j-1}^{n}-u_{i,j}^{n})}{\Delta {{y}^{2}}}
\end{split}
$%
}
\label{FiniteDifferenceApproximations}
\end{equation}
Rearrange to solve for $u_{i,j}^{n+1}$:

\begin{equation}
\resizebox{1.0\linewidth}{!}{%
$\displaystyle
\begin{split}
u_{i,j}^{n+1} = u_{i,j}^{n} + \Delta t \bigg( & \frac{{{\alpha }_{x,1}}(u_{i+1,j}^{n}-u_{i,j}^{n}) + {{\alpha }_{x,2}}(u_{i-1,j}^{n}-u_{i,j}^{n})}{\Delta {{x}^{2}}} \\
& + \frac{{{\alpha }_{y,1}}(u_{i,j+1}^{n}-u_{i,j}^{n}) + {{\alpha }_{y,2}}(u_{i,j-1}^{n}-u_{i,j}^{n})}{\Delta {{y}^{2}}} \bigg)
\end{split}
$%
}
\label{u-recursion}
\end{equation}

\begin{figure}[t]
\centering
\includegraphics[width=0.95\linewidth]{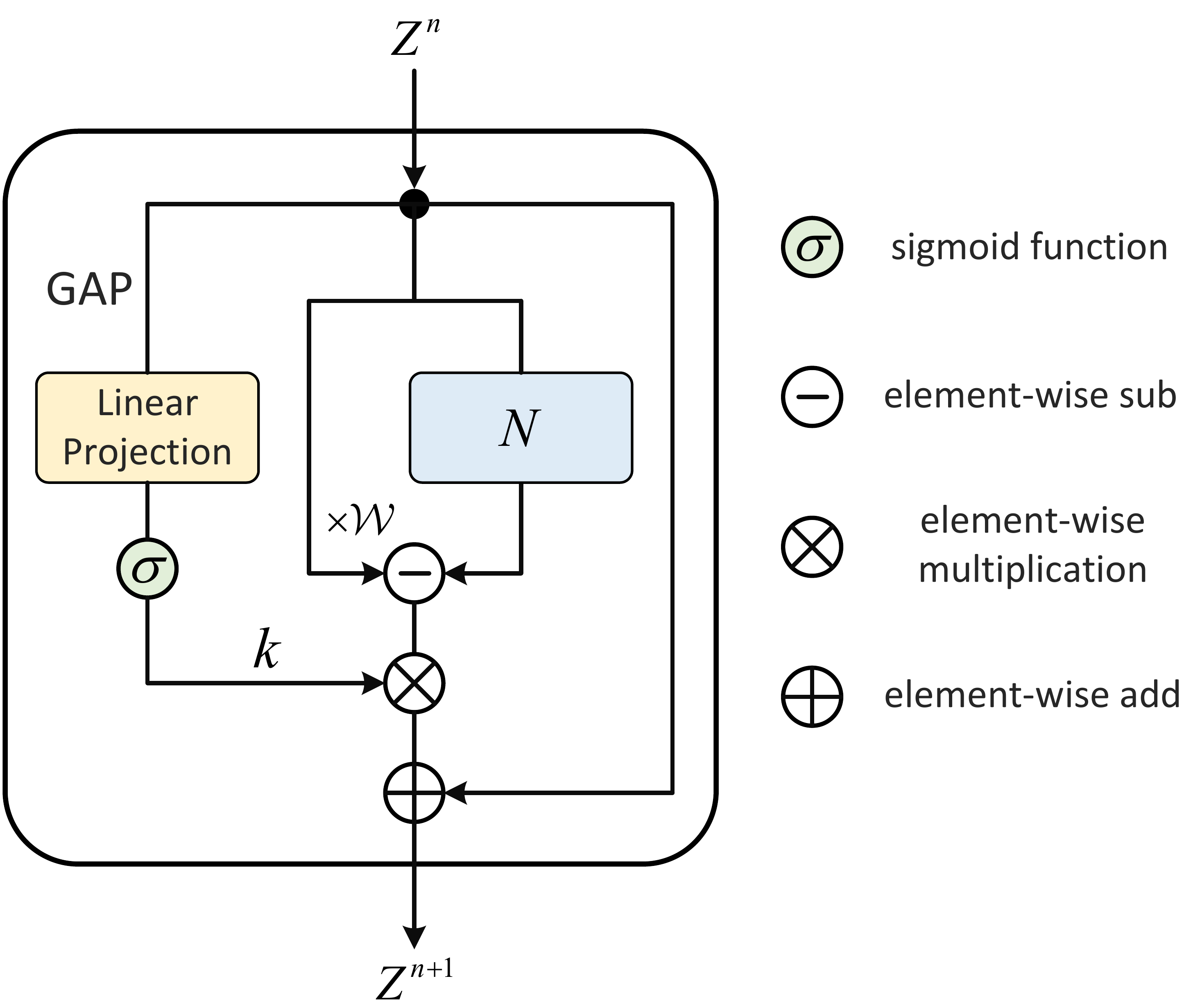} % Reduce the figure size so that it is slightly narrower than the column.
\caption{Heat Conduction Layer. $N$ denotes calculating the weighted sum of neighboring points.}
\label{HcLayer-flabel}
\end{figure}

\subsection{Heat Conduction Layer}
Suppose $Z\in {{\mathbb{R}}^{C\times H\times W}}$ is the image feature map in networks, $C$ is the number of channels, $H$ is the height of the feature, and $W$ is the width of the feature. We draw an analogy between feature information propagation in feature maps and temperature conduction, treating each channel as a separate 2-D heat conduction process, denoted as $z\in {{\mathbb{R}}^{H\times W}}$. The heat conduction layer embodies this concept of information conduction through temperature. Based on this idea, we first replace temperature $u$ in the original heat equation with feature $z$, and further set the thermal diffusivity ${{\alpha }}$ as a learnable parameter, so that the model can adaptively adjust the diffusion behavior according to the local structure of the feature map. Since the model works with discrete image patch features, it naturally leads to a grid of points where $\Delta x=\Delta y=\Delta d$. Substituting into Eq.\ref{u-recursion} yields:
\begin{equation}
\resizebox{1.0\linewidth}{!}{%
$\displaystyle
\begin{split}
z_{i,j}^{n+1} = z_{i,j}^{n} + \frac{\Delta t}{\Delta d^{2}} \bigg( & {{w}_{1}}z_{i+1,j}^{n} + {{w}_{2}}z_{i-1,j}^{n} + {{w}_{3}}z_{i,j+1}^{n} + {{w}_{4}}z_{i,j-1}^{n} \\
& - ({{w}_{1}} + {{w}_{2}} + {{w}_{3}} + {{w}_{4}})z_{i,j}^{n} \bigg)
\end{split}
$%
}
\label{feature-recursion-1}
\end{equation}
where ${{w}_{1}}={{\alpha }_{x,1}},{{w}_{2}}={{\alpha }_{x,2}},{{w}_{3}}={{\alpha }_{y,1}},{{w}_{4}}={{\alpha }_{y,2}}$, ${{w}_{i}}$ is a learnable parameter. Eq.\ref{feature-recursion-1} is further simplified as follows: 
\begin{equation}
\begin{matrix}
   z_{i,j}^{n+1}=  z_{i,j}^{n}+k\left( N\left( z_{i,j}^{n} \right)-\mathcal{W}z_{i,j}^{n} \right), & k=\frac{\Delta t}{\Delta {{d}^{2}}}  \\
\end{matrix}
\label{feature-recursion-2}
\end{equation}
where $N\left( z_{i,j}^{n} \right)$ denotes calculating the weighted sum of the neighboring points of $z_{i,j}^{n}$, $\mathcal{W}={{w}_{1}}+{{w}_{2}}+{{w}_{3}}+{{w}_{4}}$. Based on Eq.\ref{feature-recursion-2}, we propose the Heat Conduction Layer, which considers the layer dimension of the model as the time step, i.e., the feature passes through a Heat Conduction Layer and ${{t}^{n}}$ becomes ${{t}^{n+1}}$. Meanwhile, inspired by Mamba \cite{gu2024mambalineartimesequencemodeling}, we make $k$ input-dependent to increase the context-aware ability of the model. Heat Conduction Layer:

\begin{equation}
\begin{aligned}
  {{k}^{n}} = & \hspace{3pt} \sigma \left( \text{Linear}\left( \text{GAP}\left( {{Z}^{n}} \right) \right) \right) \\ 
  {{h}^{n}} = & \hspace{3pt}  N\left( {{z}^{n}} \right)-{\mathcal{W}{z}^{n}} \\ 
  {{z}^{n+1}} = & \hspace{3pt} {{z}^{n}}+{{k}^{n}}{{h}^{n}}  
\end{aligned}
\label{all-HCLayer}
\end{equation}
where ${{Z}^{n}}\in {{\mathbb{R}}^{C\times H\times W}}$, $\text{GAP:}{{\mathbb{R}}^{C\times H\times W}}\mapsto {{\mathbb{R}}^{C}}$ denotes global average pooling, $\text{Linear}$ denotes the mapping functions like fully connected layer, and $k$ is a scalar. $\sigma $ is a sigmoid function, it ensures $k>0$ and limits the size of $k$, thereby improving the stability of the FDM solution. From the principles of heat conduction, we naturally derive the residual structure \cite{He-2016-CVPR} and the SE block \cite{Hu-2018-CVPR}, used in modern model architectures. The overall structure of the Heat Conduction Layer is shown in Figure \ref{HcLayer-flabel}.

Eq.\ref{all-HCLayer} illustrates the calculation process for a single channel in the feature map ${{Z}^{n+1}}$, and the other channels are calculated in the same way.

\subsection{Preliminaries: Fourier Series}

\paragraph{Principle of Superposition.} Linear partial differential equations follow the superposition principle. For the heat conduction equation, this principle states that if ${{u}_{1}}\left( x,y,t \right),{{u}_{2}}\left( x,y,t \right),\cdots ,{{u}_{n}}\left( x,y,t \right)$ are both solutions to the heat equation, any linear combination of these solutions is also a solution:

\begin{equation}
u\left( x,y,t \right)=\sum\nolimits_{i=1}^{n}{{{a}_{i}}{{u}_{i}}\left( x,y,t \right)}
\label{SuperpositionPrinciple}
\end{equation}
where ${{u}_{1}},{{u}_{2}},\cdots ,{{u}_{n}}$ satisfy Eq.\ref{HeatEquation}, $u\left( x,y,t \right)$ is also a solution for any constants ${{a}_{1}},{{a}_{2}},\cdots ,{{a}_{n}}$. The principle of superposition allows us to construct complex solutions to the heat equation from simpler ones.

\paragraph{Fourier Series.} The Fourier series is a powerful tool for solving the heat equation. The Fourier series solution of the heat equation in a two-dimensional rectangular space is as follows:

\begin{equation}
\resizebox{\linewidth}{!}{
$u\left( x,y,t \right)=\sum\limits_{m=1}^{M}{\sum\limits_{n=1}^{N}{{{B}_{mn}}\sin \left( \frac{m\pi x}{L} \right)\sin \left( \frac{n\pi y}{L} \right){{e}^{-\frac{k\left( {{m}^{2}}+{{n}^{2}} \right){{\pi }^{2}}t}{{{L}^{2}}}}}}}$
}
\label{FourierSeries}
\end{equation}
where $m$ and $n$ denote constant frequencies, ${{B}_{mn}}$ is the Fourier coefficient, and $L$ is the length of the sides of the rectangular region. Further simplification of Eq.\ref{FourierSeries}: 

\begin{equation}
u\left( x,y,t \right)=\sum\limits_{m=1}^{M}{{{\beta }_{m}}{{g}_{m}}\left( x,y,t \right)}
\label{FourierSeries-S}
\end{equation}
where ${{g}_{m}}\left( x,y,t \right)=\sin \left( \frac{m\pi x}{L} \right)\sin \left( \frac{n\pi y}{L} \right){{e}^{-\frac{k\left( {{m}^{2}}+{{n}^{2}} \right){{\pi }^{2}}t}{{{L}^{2}}}}}$, ${{\beta }_{m}}={{B}_{mn}}$, ${{\beta }_{m}}$ is a constant. From Eq.\ref{FourierSeries-S}, we can conclude that the Fourier series fits the complex heat conduction process by linear combination of basis functions $\left\{ {{g}_{m}}\left( x,y,t \right) \right\}_{m=1}^{M}$ at different frequencies, and becomes a more precise approximation as more terms are included. That is, as $M$ increases, it can represent more intricate patterns and capture much finer details.

\begin{figure}[t]
\centering
\includegraphics[width=0.75\linewidth]{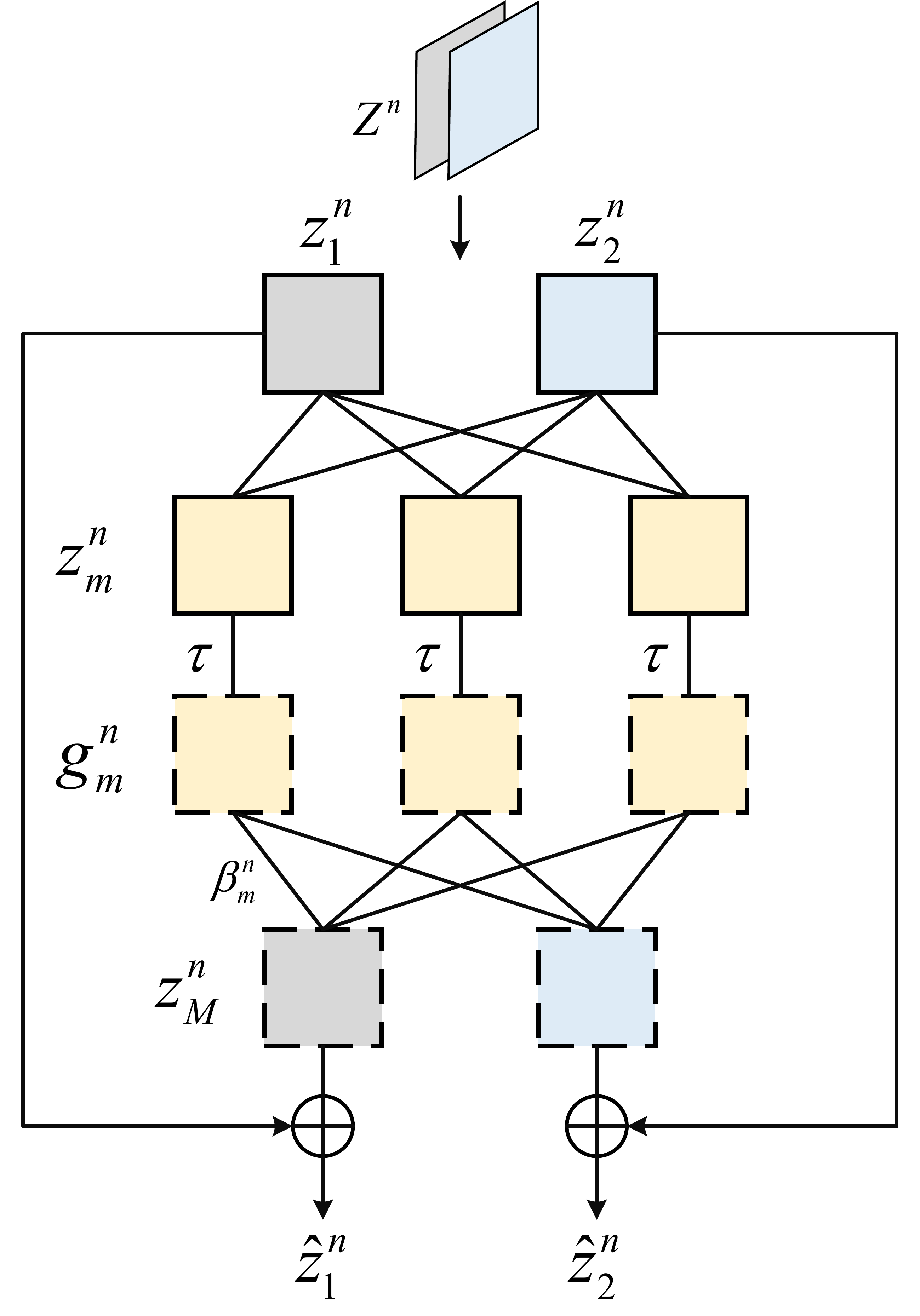} % Reduce the figure size so that it is slightly narrower than the column.
\caption{Refinement Approximation Layer. We take a feature map with two channels as an example; feature maps with more channels are processed in a similar manner.}
\label{RaLayer-flabel}
\end{figure}

\subsection{Refinement Approximation Layer}

In \emph{Heat Conduction Layer} subsection, we have illustrated the heat conduction of features on a single channel. To effectively extract visual features, it is also necessary to integrate information from multiple channels. We consider each channel as a separate 2-D heat conduction process. Therefore, each channel can be approximated using the Fourier series. According to Eq.\ref{FourierSeries-S}:

\begin{equation}
{{z}_{c}}\left( x,y,{{t}^{n}} \right)=\sum\limits_{h=1}^{H}{{{\beta }_{h}}{{g}_{h}}\left( x,y,{{t}^{n}} \right)}
\label{Feature-FourierSeries-S}
\end{equation}
where ${{z}_{c}}\left( x,y,{{t}^{n}} \right)$ denotes the featured heat distribution of the $c$-th channel of the ${{t}^{n}}$ time-step. The accuracy of the approximation can be improved by increasing the number of terms in the Fourier series:

\begin{equation}
\begin{aligned}
\hat{z}_{c}\left( x,y,{t^{n}} \right) &= \sum_{h=1}^{H} \beta_{h} g_{h}\left( x,y,{t^{n}} \right) + \sum_{m=H}^{M} \beta_{m} g_{m}\left( x,y,{t^{n}} \right) \\
&= z_{c}\left( x,y,{t^{n}} \right) + \sum_{m=H}^{M} \beta_{m} g_{m}\left( x,y,{t^{n}} \right)
\end{aligned}
\label{add-FourierSeries-S}
\end{equation}

Based on Eq.\ref{add-FourierSeries-S}, we propose the Refinement Approximation Layer. Mathematically, considering the input as ${{Z}^{n}}=\left\{ {{z}_{c}}\left( x,y,{{t}^{n}} \right) \right\}_{c=1}^{C}$. Firstly, based on Eq.\ref{SuperpositionPrinciple}, we use $\left\{ {{z}_{c}}\left( x,y,{{t}^{n}} \right) \right\}_{c=1}^{C}$ to construct more new solutions as candidate terms $\left\{ {{z}_{m}}\left( x,y,{{t}^{n}} \right) \right\}_{m=1}^{M-H}$. It is worth noting that the simple linear combination cannot generate new frequencies, so we perform a nonlinear transformation of $\left\{ {{z}_{m}}\left( x,y,{{t}^{n}} \right) \right\}_{m=1}^{M-H}$ to introduce new frequencies to obtain new basis functions $\left\{ {{g}_{m}} \right\}_{m=1}^{M-H}$. This can be done by a nonlinear activation function, taking the GELU \cite{hendrycks2023gaussianerrorlinearunits} as an example:
\begin{equation}
\text{GELU}\left( z \right)\approx 0.5z\left( 1+\tanh \left( \sqrt{\frac{2}{\pi }}\left( z+0.044715{{z}^{3}} \right) \right) \right)
\label{GELU}
\end{equation}
where higher-order terms (${{z}^{2}}$, ${{z}^{3}}$) will bring new frequencies, take ${{z}^{2}}$ as an example:
\begin{equation} 
\resizebox{\linewidth}{!}{
$\begin{aligned} z\left( x,y,{t^{n}} \right)^{2} &= B_{mn}^{2} \sin^{2}\left( \frac{m\pi x}{L} \right) \sin^{2}\left( \frac{n\pi y}{L} \right) e^{-2\frac{k\left( m^{2}+n^{2} \right)\pi^{2} t^{n}}{L^{2}}} \\ &= \frac{B_{mn}^{2}}{4} e^{-2\frac{k(m^{2}+n^{2})\pi^{2} t^{n}}{L^{2}}} \Bigg( 1 - \cos \left( \frac{2m\pi x}{L} \right) - \cos \left( \frac{2n\pi y}{L} \right) \\ & \quad + \frac{1}{2} \cos \left( \frac{2m\pi x}{L} + \frac{2n\pi y}{L} \right) + \frac{1}{2} \cos \left( \frac{2m\pi x}{L} - \frac{2n\pi y}{L} \right) \Bigg) \end{aligned}$
}
\label{higher-order-term}
\end{equation}

\begin{figure*}[t]
\centering
\includegraphics[width=0.93\linewidth]{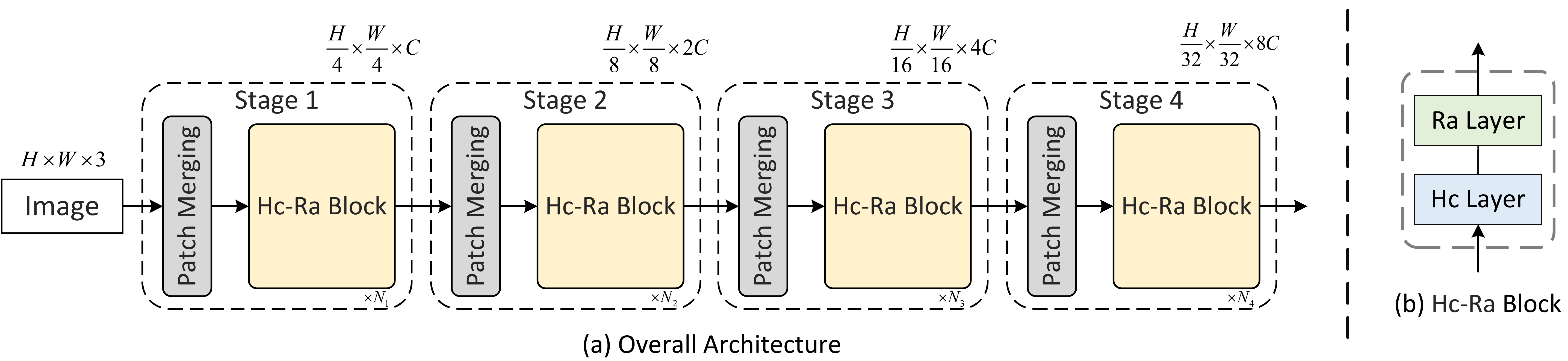} % Reduce the figure size so that it is slightly narrower than the column.
\caption{\textbf{(a) The overall framework of HcNet.} HcNet adopts hierarchical architecture with 4 stages, where stage [1, 2, 3, 4] have [${{N}_{1}}$, ${{N}_{2}}$, ${{N}_{3}}$, ${{N}_{4}}$] blocks, respectively. \textbf{(b) The architecture of Hc-Ra block.} Compared with transformer block, it does not draw the residual structure, as it is inherently incorporated within the heat conduction layer or the refinement approximation layer.}
\label{OverallArchitecture-flabel}
\end{figure*}

New frequencies, such as $2m$ and $2n$, are generated in Eq.\ref{higher-order-term}. Although nonlinear transformations can generate new higher-order harmonics, they also introduce undesirable components (e.g., DC component) and make it challenging to precisely control the amplitude of specific Fourier series terms. To address this issue, we introduce convolution operations as filters, enabling the model to adaptively select Fourier series terms with specific frequencies and amplitudes. Specifically, we input the results of the nonlinear transformation into a depth-wise separable convolution to obtain the final basis functions $\left\{ {{g}_{m}} \right\}_{m=1}^{M-H}$. Extensive research has demonstrated that convolutional kernels can be regarded as filters \cite{NIPS2014-81ca0262,Zoumpourlis-2017-ICCV}. Then, based on Eq.\ref{add-FourierSeries-S}, each channel uses a separate set of ${{\beta }_{m}}$ to linearly combine ${{g}_{m}}$ to obtain the new term $\sum\limits_{m=H}^{M}{{{\beta }_{m}}{{g}_{m}}\left( x,y,{{t}^{n}} \right)}$. Finally, add the new term to the input ${{z}_{c}}\left( x,y,{{t}^{n}} \right)$ to get ${{\hat{z}}_{c}}\left( x,y,{{t}^{n}} \right)$, where the number of terms increases from $H$ to $M$. Refinement Approximation Layer:

\begin{equation}
\begin{aligned}
  z_{m}^{n}= & \hspace{3pt} \text{Linear}\left( {{Z}^{n}} \right) \\ 
 g_{m}^{n}= & \hspace{3pt} \tau \left( z_{m}^{n} \right) \\ 
 z_{M}^{n}= & \hspace{3pt} \text{Linear}\left( g_{m}^{n} \right) \\ 
 {{\hat{z}}^{n}}= & \hspace{3pt} {{z}^{n}}+z_{M}^{n}  
\end{aligned}
\label{RA-Layer}
\end{equation}
where ${{Z}^{n}}$ denotes the feature at the $n$-th time-step. $\tau \left( \cdot  \right)$ denotes the nonlinear transformation, which consists of a nonlinear activation function and a depth-wise convolutional layer. $z_{M}^{n}$ denotes the new term $\sum\limits_{m=H}^{M}{{{\beta }_{m}}{{g}_{m}}\left( x,y,{{t}_{n}} \right)}$, and the overall flow is shown in Figure \ref{RaLayer-flabel}. The Refinement Approximation Layer increases the number of Fourier series terms in each channel, so as the depth of the model increases, the model's ability to capture complex patterns and dynamics becomes stronger.

From the perspective of the heat equation, the role of the nonlinear activation function is to generate Fourier series terms with new frequencies, while the role of the MLP layer is to linearly combine the basis functions to approximate any function.

\begin{table}[h]
\centering
\renewcommand{\arraystretch}{0.9}
\small
\resizebox{\linewidth}{!}{
\begin{tabular}{c|*{4}{c}|*{4}{c}} %
\toprule
\multirow{2}{*}{Version} & \multicolumn{4}{c|}{Blocks} & \multicolumn{4}{c}{Dims} \\
% \cmidrule(r){2-5} \cmidrule(l){6-9} % 
 & $S_1$ & $S_2$ & $S_3$ & $S_4$ & $D_1$ & $D_2$ & $D_3$ & $D_4$ \\
\midrule
HcNet-T  & 5 & 5 & 14  & 5 &  64 & 128 & 320 & 512 \\
HcNet-S & 6 & 6 & 18  & 6 &  64 & 192 & 384 & 768 \\
HcNet-B  & 10 & 10 & 28 & 10 &  96 & 192 & 384 & 768 \\
\bottomrule
\end{tabular}
}
\caption{Detailed settings of HcNet series.}
\label{HcNetArchitecture}
\end{table}

\begin{table}[t]
   \centering
	\renewcommand\arraystretch{1.3}
	\Huge
	\resizebox{\linewidth}{!}{
   \begin{tabular}{c|lcc|c}
       \toprule
        General Arch. &Model   & Params & MACs                & Top-1 acc.           \\
       \midrule
	\multirow{5}{*}{CNNs} 
       &RegNetY-4G \cite{Radosavovic-2020-CVPR}  & 21M   & 4.0G     &80.0    \\
	 &RegNetY-16G \cite{Radosavovic-2020-CVPR}  & 84M   & 16.0G     &82.9    \\
      &ConvNeXt-T \cite{Liu-2022-CVPR}  & 29M   & 4.5G     &82.1    \\
	&ConvNeXt-S \cite{Liu-2022-CVPR}  & 50M   & 8.7G     &83.1    \\
	&ConvNeXt-B \cite{Liu-2022-CVPR}  & 89M   & 15.4G     &83.8    \\
	 \midrule
	\multirow{13}{*}{ViTs} 
	  &ViT-B  \cite{DBLP:journals/corr/abs-2010-11929}  & 86M   & 55.4G     &77.9     \\
	  &Swin-T \cite{Liu-2021-ICCV}  & 28M   & 4.6G     &81.3    \\
       &Swin-S \cite{Liu-2021-ICCV}  & 50M   & 8.7G     &83.0    \\
	 &Swin-B \cite{Liu-2021-ICCV}  & 88M   & 15.4G     &83.5    \\
	 &BiFormer-T \cite{Zhu-2023-CVPR}  & 13M   & 2.2G     &81.4    \\
	 &BiFormer-S \cite{Zhu-2023-CVPR}  & 26M   & 4.5G     &83.8    \\
	 &BiFormer-B \cite{Zhu-2023-CVPR}  & 57M   & 9.8G     &84.3    \\
	 &CrossFormer-S \cite{10366193}  & 31M   & 4.9G     &82.5    \\
	 &CrossFormer-B \cite{10366193}  & 52M   & 9.2G     &83.4    \\
	 &CrossFormer-L \cite{10366193}  & 92M   & 16.1G     &84.0    \\
 	 &QFormer-T \cite{10384565}  & 29M   & 4.6G     &82.5    \\
	 &QFormer-S \cite{10384565}  & 51M   & 8.9G     &84.0    \\
	 &QFormer-B \cite{10384565}  & 90M   & 15.7G     &84.1    \\
	 \midrule
	\multirow{8}{*}{PIMs} 
	  &Vim-S \cite{pmlr-v235-zhu24f}  & 26M   & 5.3G     &81.4    \\
       &Vim-B \cite{pmlr-v235-zhu24f}  & 98M   & 19.0G     &83.2    \\
	&VMamba-T \cite{liu2024vmambavisualstatespace}  & 31M   & 4.9G     &82.6    \\
	&VMamba-S \cite{liu2024vmambavisualstatespace}  & 50M   & 8.7G     &83.6    \\
	&VMamba-B \cite{liu2024vmambavisualstatespace}  & 89M   & 15.4G     &83.9    \\
	&\multicolumn{1}{>{\columncolor{gray!40}}l}{HcNet-T }& \multicolumn{1}{>{\columncolor{gray!40}}c}{28M}& \multicolumn{1}{>{\columncolor{gray!40}}c|}{4.1G} &\multicolumn{1}{>{\columncolor{gray!40}}c}{83.0} \\
	&\multicolumn{1}{>{\columncolor{gray!40}}l}{HcNet-S }& \multicolumn{1}{>{\columncolor{gray!40}}c}{51M}& \multicolumn{1}{>{\columncolor{gray!40}}c|}{8.6G} &\multicolumn{1}{>{\columncolor{gray!40}}c}{83.9} \\
	&\multicolumn{1}{>{\columncolor{gray!40}}l}{HcNet-B }& \multicolumn{1}{>{\columncolor{gray!40}}c}{85M}& \multicolumn{1}{>{\columncolor{gray!40}}c|}{14.2G} &\multicolumn{1}{>{\columncolor{gray!40}}c}{84.3} \\
       \bottomrule
   	\end{tabular}
	}
   \caption{Comparison of different types of models on ImageNet-1K. \textbf{PIMs} denote physics inspired models.}
   \label{ImageNet-Top1}
\end{table}

\begin{table*}[t]
\centering
\renewcommand\arraystretch{1.2}
\large
\resizebox{\linewidth}{!}{
\begin{tabular}{c|c|cc|cccccc|cccccc}
\toprule
\multicolumn{2}{c|}{\multirow{2}{*}{Backbone}} &\multirow{2}{*}{Params}&\multirow{2}{*}{MACs} & \multicolumn{6}{c|}{Mask R-CNN 1× schedule}&\multicolumn{6}{c}{Mask R-CNN 3× schedule + MS}\\

\multicolumn{2}{c|}{} & & & $\text{AP}^{\text{box}}$ &  $\text{AP}^{\text{box}}_{\text{50}}$  &  $\text{AP}^{\text{box}}_{\text{75}}$ &  $\text{AP}^{\text{mask}}$ &  $\text{AP}^{\text{mask}}_{\text{50}}$ &  $\text{AP}^{\text{mask}}_{\text{75}}$  & $\text{AP}^{\text{box}}$ &  $\text{AP}^{\text{box}}_{\text{50}}$  &  $\text{AP}^{\text{box}}_{\text{75}}$ &  $\text{AP}^{\text{mask}}$ &  $\text{AP}^{\text{mask}}_{\text{50}}$ &  $\text{AP}^{\text{mask}}_{\text{75}}$        \\
\midrule

\multicolumn{2}{c|}{ResNet-50 \cite{He-2016-CVPR}} & 44M   & 260G   &38.0  &58.6 &41.4&34.4&55.1&36.7 &41.0& 61.7& 44.9& 37.1& 58.4 &40.1\\ 
\multicolumn{2}{c|}{Swin-T \cite{Liu-2021-ICCV}} & 48M   & 264G   &43.7&66.6 &47.6&39.8&63.3&42.7 &46.0&68.2&50.2&41.6&65.1&44.8\\ 
\multicolumn{2}{c|}{ Focal-T \cite{DBLP:journals/corr/abs-2107-00641}} & 49M   & 291G   &44.8&67.7 &49.2&41.0&64.7&44.2 &47.1&69.4&51.9&42.7&66.5&45.9\\
\multicolumn{2}{c|}{ CrossFormer-S \cite{10366193}} & 50M   & 301G   &45.4&68.0 &49.7&41.4&64.8&44.6 &--&--&--&--&--&--\\
\multicolumn{2}{c|}{ QFormer-T \cite{10384565}} & 49M   & --   &45.9&68.5 &50.3&41.5&65.2&44.6 &47.5&69.6&52.1&42.7&66.4&46.1\\
\multicolumn{2}{c|}{ VMamba-T \cite{liu2024vmambavisualstatespace}} & 42M   & 262G   &46.5&68.5 &50.7&42.1&65.5&45.3 &48.5&69.9&52.9&43.2&66.8&46.3\\ 
\multicolumn{2}{>{\columncolor{gray!40}}c|}{HcNet-T} & \multicolumn{1}{>{\columncolor{gray!40}}c}{35M}  &\multicolumn{1}{>{\columncolor{gray!40}}c|}{242G}&\multicolumn{1}{>{\columncolor{gray!40}}c}{46.0}&\multicolumn{1}{>{\columncolor{gray!40}}c}{68.3} &\multicolumn{1}{>{\columncolor{gray!40}}c}{50.2}&\multicolumn{1}{>{\columncolor{gray!40}}c}{41.9}&\multicolumn{1}{>{\columncolor{gray!40}}c}{65.4}&\multicolumn{1}{>{\columncolor{gray!40}}c|}{44.7} &\multicolumn{1}{>{\columncolor{gray!40}}c}{47.6}&\multicolumn{1}{>{\columncolor{gray!40}}c}{69.3}&\multicolumn{1}{>{\columncolor{gray!40}}c}{52.3}&\multicolumn{1}{>{\columncolor{gray!40}}c}{42.8}&\multicolumn{1}{>{\columncolor{gray!40}}c}{66.5}&\multicolumn{1}{>{\columncolor{gray!40}}c}{45.8}\\ 

\midrule

\multicolumn{2}{c|}{ResNet-101 \cite{He-2016-CVPR}} & 63M   & 336G   &40.4  &61.1 &44.2&36.4&57.7&38.8 &42.8&63.2&47.1&38.5&60.1&41.3\\ 
\multicolumn{2}{c|}{Swin-S \cite{Liu-2021-ICCV}} & 69M   & 354G   &44.8&66.6 &48.9&40.9&63.2&44.2 &48.2&69.8&52.8&43.2&67.0&46.1\\ 
\multicolumn{2}{c|}{CSWin-S \cite{Dong-2022-CVPR}} & 54M   & 342G   &47.9&70.1 &52.6&43.2&67.1&46.2 &50.0&71.3&54.7&44.5&68.4&47.7\\ 
\multicolumn{2}{c|}{CrossFormer-B \cite{10366193}} & 71M   & 408G   &47.2&69.9 &51.8&42.7&66.6&46.2 &--&--&--&--&--&--\\ 
\multicolumn{2}{c|}{QFormer-S \cite{10384565}} & 70M   & --   &--&-- &--&--&--&-- &49.5&71.2&54.2&44.2&68.3&47.6\\ 
\multicolumn{2}{c|}{ VMamba-S \cite{liu2024vmambavisualstatespace}} & 64M   & 357G   &48.2&69.7 &52.5&43.0&66.6&46.4 &49.7&70.4&54.2&44.0&67.6&47.3\\ 
\multicolumn{2}{>{\columncolor{gray!40}}c|}{HcNet-S} & \multicolumn{1}{>{\columncolor{gray!40}}c}{50M}  &\multicolumn{1}{>{\columncolor{gray!40}}c|}{311G}&\multicolumn{1}{>{\columncolor{gray!40}}c}{47.6}&\multicolumn{1}{>{\columncolor{gray!40}}c}{69.3} &\multicolumn{1}{>{\columncolor{gray!40}}c}{52.2}&\multicolumn{1}{>{\columncolor{gray!40}}c}{43.1}&\multicolumn{1}{>{\columncolor{gray!40}}c}{66.5}&\multicolumn{1}{>{\columncolor{gray!40}}c|}{46.3} &\multicolumn{1}{>{\columncolor{gray!40}}c}{49.8}&\multicolumn{1}{>{\columncolor{gray!40}}c}{71.1}&\multicolumn{1}{>{\columncolor{gray!40}}c}{54.6}&\multicolumn{1}{>{\columncolor{gray!40}}c}{44.7}&\multicolumn{1}{>{\columncolor{gray!40}}c}{68.1}&\multicolumn{1}{>{\columncolor{gray!40}}c}{47.7}\\ 

\midrule

\multicolumn{2}{c|}{RegNeXt-101-64 \cite{Xie-2017-CVPR}} & 101M   & 493G   &42.8  &63.8 &47.3&38.4&60.6&41.3 &44.4&64.9&48.8&39.7&61.9&42.6\\ 
\multicolumn{2}{c|}{Swin-B \cite{Liu-2021-ICCV}} & 107M   & 496G   &46.9&-- &--&42.3&--&-- &48.5&69.8&53.2&43.4&66.8&46.9\\ 
\multicolumn{2}{c|}{CSWin-B \cite{Dong-2022-CVPR}} & 97M   & 526G   &48.7&70.4 &53.9&43.9&67.8&47.3 &50.8&72.1&55.8&44.9&69.1&48.3\\ 
\multicolumn{2}{c|}{VMamba-B \cite{liu2024vmambavisualstatespace}} & 108M   & 485G   &49.2&70.9 &53.9&43.9&67.7&47.6 &--&--&--&--&--&--\\ 
\multicolumn{2}{>{\columncolor{gray!40}}c|}{HcNet-B} & \multicolumn{1}{>{\columncolor{gray!40}}c}{90M}  &\multicolumn{1}{>{\columncolor{gray!40}}c|}{473G}&\multicolumn{1}{>{\columncolor{gray!40}}c}{48.8}&\multicolumn{1}{>{\columncolor{gray!40}}c}{70.2} &\multicolumn{1}{>{\columncolor{gray!40}}c}{53.7}&\multicolumn{1}{>{\columncolor{gray!40}}c}{44.3}&\multicolumn{1}{>{\columncolor{gray!40}}c}{67.9}&\multicolumn{1}{>{\columncolor{gray!40}}c|}{47.5} &\multicolumn{1}{>{\columncolor{gray!40}}c}{50.4}&\multicolumn{1}{>{\columncolor{gray!40}}c}{71.9}&\multicolumn{1}{>{\columncolor{gray!40}}c}{55.6}&\multicolumn{1}{>{\columncolor{gray!40}}c}{45.0}&\multicolumn{1}{>{\columncolor{gray!40}}c}{69.3}&\multicolumn{1}{>{\columncolor{gray!40}}c}{48.5}\\ 

\bottomrule

\end{tabular}
}
\caption{Object detection and instance segmentation performance on the COCO val2017 with the Mask R-CNN framework. The models have been pre-trained on ImageNet-1K. The resolution used to calculate MACs is 800×1280.}
\label{COCO-Top1}
\end{table*}

\subsection{HcNet Model}

Equipped with the above Heat Conduction layer and Refinement Approximation layer, the Hc-Ra block is shown in Figure \ref{OverallArchitecture-flabel} (b). We take advantage of the Hc-Ra block by adopting a hierarchical structure similar to hierarchical transformer variants \cite{Liu-2021-ICCV,Zhu-2023-CVPR}. Figure \ref{OverallArchitecture-flabel} (a) shows the overall framework of HcNet. Specifically, HcNet has 4 stages, each stage contains a patch merging layer and several Hc-Ra blocks. As the network gets deeper, the input features are spatially downsampled by a certain ratio through the patch merging layer. The channel dimension is expanded twice to produce a hierarchical image representation. Specifically, the spatial downsampling ratio is set to 4 in the first stage and 2 in the last three stages. The outputs of the patch merging layer are fed into the subsequent Hc-Ra block, and the number of tokens is kept constant. 

Based on this architecture, we build three different variants of HcNet, including HcNet-T, HcNet-S, and HcNet-B, as shown in Table \ref{HcNetArchitecture}.

\section{Experiments}

To demonstrate the effectiveness of the HcNet, we conduct experiments on ImageNet-1K \cite{5206848}. To further evaluate the generalization and robustness of our backbone, we also conduct experiments on ADE20K \cite{Zhou-2017-CVPR} for semantic segmentation, and COCO \cite{10.1007/978-3-319-10602-1-48} for object detection. Finally, we perform comprehensive ablation studies to analyze each component of the HcNet.

\subsection{Classiﬁcation on the ImageNet-1K}

\paragraph{Implementation details.} This setting mostly follows Swin \cite{Liu-2021-ICCV}. We use the PyTorch toolbox \cite{paszke2019pytorch} to implement all our experiments. We employ an AdamW \cite{kingma2014adam} optimizer for 300 epochs using a cosine decay learning rate scheduler and 20 epochs of linear warm-up. A batch size of 256, an initial learning rate of 0.001, and a weight decay of 0.05 are used. ViT-B/16 uses an image size 384×384 and others use 224×224. We include most of the augmentation and regularization strategies of Swin transformer \cite{Liu-2021-ICCV} in training.

\paragraph{Results.} Table \ref{ImageNet-Top1} compares the performance of the proposed HcNet with the state-of-the-art CNN, Vision Transformer and PIM backbones on ImageNet-1K. Surprisingly, despite the simple heat conduction layer, HcNets can still achieve highly competitive performance compared with CNNs, ViTs and PIMs. For example, HcNet-T reaches the top-1 accuracy of 83.0 while only requiring 28M parameters and 4.1G MACs. Compared with ViT-B and Swin-T, HcNet-T outperforms them by 5.1 and 1.7, respectively. With similar parameters and MACs, HcNet performs better than other PIMs. Experimental results show that physics inspired models have the potential to achieve reasonable performance, and their interpretability is unmatched by other models.

\begin{figure*}[t]
\centering
\includegraphics[width=0.98\linewidth]{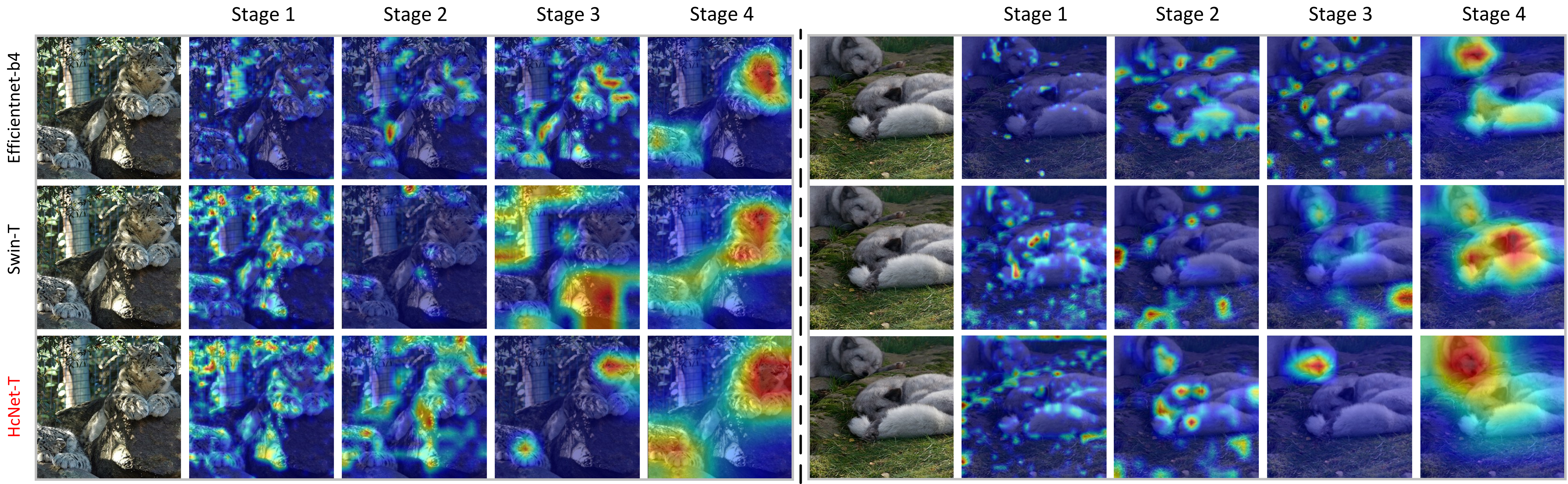} % Reduce the figure size so that it is slightly narrower than the column.
\caption{Grad-CAM visualization of the models trained on ImageNet-1K. The visualized images are randomly selected from the test dataset.}
\label{CamMainText-flabel}
\end{figure*}

\subsection{Object Detection and Instance Segmentation}

\paragraph{Implementation details.} We use the Mask R-CNN \cite{He-2017-ICCV} framework to evaluate the performance of the proposed HcNet backbone on the COCO benchmark for object detection and instance segmentation. We pretrain the backbones on the ImageNet-1K dataset and apply the ﬁnetuning strategy used in Swin Transformer \cite{Liu-2021-ICCV} on the COCO training set.

\paragraph{Results.} Table \ref{COCO-Top1} reports the results of the Mask R-CNN framework with “1×” (12 training epoch) and “3 × +MS” (36 training epoch with multi-scale training) schedule. With a 12-epoch fine-tuning schedule, HcNet-T/S/B models achieve object detection mAPs of 46.0\%/47.6\%/48.8\%, outperforming Swin-T/S/B by 2.3\%/2.8\%/1.9\% bAP, respectively. HcNets have similar performance in instance segmentation. Overall, for COCO object detection and instance segmentation, HcNets achieve competitive performance.

\subsection{Semantic Segmentation}

\paragraph{Implementation details.} We further investigate the capability of HcNet for Semantic Segmentation on the ADE20K \cite{Zhou-2017-CVPR} dataset. Here we employ the widely-used UperNet \cite{Xiao-2018-ECCV} as the basic framework and followed Swin's \cite{Liu-2021-ICCV} experimental settings. In Table \ref{ADE20K-Top1}, we report both the single-scale (SS) and multi-scale (MS) mIoU for better comparison. The default input resolution is 512 × 512.

\paragraph{Results.} As shown in Table \ref{ADE20K-Top1}, HcNet can attain competitive performance on semantic segmentation. This further indicates the great potential of physics inspired models.

\begin{table}[t]
   \centering
   \renewcommand\arraystretch{1.3}
   \large
   \resizebox{\linewidth}{!}{
   \begin{tabular}{c|cc|cc}
      \toprule
      Backbone  & Params & MACs & SS mIoU & MS mIoU     \\
       \midrule
      ResNet-50 \cite{He-2016-CVPR} & 67M   & 953G   &42.1&42.8     \\
 	ConvNeXt-T \cite{Liu-2022-CVPR} & 60M   & 939G   &46.0&46.7    \\
      Swin-T \cite{Liu-2021-ICCV} & 60M   & 945G   &44.4&45.8    \\
	CSWin-T \cite{Dong-2022-CVPR} & 60M   & 959G   &49.3&--     \\
	CrossFormer-S \cite{10366193} & 62M   & 979G   &47.6&48.4     \\
	QFormer-T \cite{10384565} & 61M   & --   &46.9&48.1     \\
	 Vim-S \cite{pmlr-v235-zhu24f} & 46M   & --   &44.9& --   \\
      VMamba-T \cite{liu2024vmambavisualstatespace} & 62M   & 948G   &48.3& 48.6   \\
	 \multicolumn{1}{>{\columncolor{gray!40}}c|}{HcNet-T} & \multicolumn{1}{>{\columncolor{gray!40}}c}{50M} & \multicolumn{1}{>{\columncolor{gray!40}}c|}{851G}&\multicolumn{1}{>{\columncolor{gray!40}}c}{48.1}& \multicolumn{1}{>{\columncolor{gray!40}}c}{48.3}   \\
      \midrule
	ResNet-101 \cite{He-2016-CVPR} & 85M   & 1030G   &42.9&44.0     \\
	ConvNeXt-B \cite{Liu-2022-CVPR} & 122M   & 1170G   &49.1 &49.9     \\
      Swin-B \cite{Liu-2021-ICCV} & 121M   & 1188G   &48.1&49.7     \\
	CSWin-B \cite{Dong-2022-CVPR} & 109M   & 1222G   &51.1&--     \\
	CrossFormer-L \cite{10366193} & 125M   & 1258G   &50.4&51.4     \\
	QFormer-B \cite{10384565} & 123M   & --   &49.5&50.6     \\
	VMamba-B \cite{liu2024vmambavisualstatespace} & 122M  &1170G   &51.0 &51.6     \\
	 \multicolumn{1}{>{\columncolor{gray!40}}c|}{HcNet-B} & \multicolumn{1}{>{\columncolor{gray!40}}c}{96M} & \multicolumn{1}{>{\columncolor{gray!40}}c|}{1095G}&\multicolumn{1}{>{\columncolor{gray!40}}c}{51.0}& \multicolumn{1}{>{\columncolor{gray!40}}c}{51.5}   \\

       \bottomrule
   \end{tabular}
   }
\caption{Comparison of the segmentation performance of different backbones on the ADE20K. All backbones are pretrained on ImageNet-1K.}
\label{ADE20K-Top1}
\end{table}

\subsection{Visualization Analysis}

To comprehensively evaluate the quality of the learned visual representations, we employ the GradCAM technique \cite{Selvaraju-2017-ICCV} to generate activation maps for visual representations at various stages of efficientnet-b4 \cite{pmlr-v97-tan19a}, Swin-T, and HcNet-T. These activation maps provide insight into the significance of individual pixels in depicting class discrimination for each input image. As shown in Figure \ref{CamMainText-flabel}, the activation maps generated by HcNet are significantly different from the other two types of models, and identifying more semantically meaningful regions in deeper layers.

\begin{table}[t]
   \centering
   \renewcommand\arraystretch{0.9}
   \small
   \begin{tabular}{lcc}
      \toprule
      Settings  & HcNet-T & HcNet-B    \\
       \midrule
      Fixed $k=0.1$  & 81.3   & 82.7     \\
	Fixed $k=1.0$  & 81.2    & 82.8     \\
	Fixed $k=3.0$  & 81.0    & 82.5    \\
	$k$ as a learnable parameter & 82.2  & 83.3   \\
 	$k$ is input-dependent &  \textbf{83.0} &  \textbf{84.3} \\
       \bottomrule
   \end{tabular}
\caption{Compare different ways of getting $k$. The table shows the Top-1 acc.}
\label{ThermalDiffusivityTimeStep}
\end{table}

\begin{table}[t]
   \centering
   \renewcommand\arraystretch{0.9}
   \small
   \begin{tabular}{lcc}
      \toprule
        & HcNet-T  & HcNet-B    \\
       \midrule
      without filter  & 82.1     & 83.5    \\
 	with filter &  \textbf{83.0} &  \textbf{84.3} \\
       \bottomrule
   \end{tabular}
\caption{Impact of filters. The table shows the Top-1 acc.}
\label{dwconv}
\end{table}

\subsection{Ablation Studies}

The experiments of ablation studies are conducted on ImageNet-1K. We discuss the ablation below according to the following aspects.

\paragraph{The way to get $k$.} $k=\frac{\Delta t}{\Delta {{d}^{2}}}$ is the ratio of the time step to the spatial step. To demonstrate the effectiveness of input-dependent, we conduct the following experiments on ImageNet-1K: (1) fix $k$, (2) treat $k$ as a learnable parameter for each layer, and (3) $k$ in each layer depends on the input. As shown in Table \ref{ThermalDiffusivityTimeStep}, setting $k$ to be input-dependent improves the performance significantly.

\paragraph{Filters.} The generated new frequency terms are filtered using depth-wise convolution within the Refinement Approximation Layer. In this subsection, we explore the impact of depth-wise convolution on the model's performance. The model performance while keeping the same number of parameters is shown in Table \ref{dwconv}, and the results prove the necessity of filtering the Fourier series terms that introduce new frequencies after nonlinear transformation.

\section{Conclusion}

In this paper, we draw a relatively comprehensive connection between modern model architectures and the heat conduction equation. Such connection enables us to design new and more interpretable neural network architectures. Benefiting from this connection, we propose the Heat Conduction Layer and the Refinement Approximation Layer inspired by solving the heat conduction equation using Finite Difference Method and Fourier series, respectively. Extensive experiments demonstrate that HcNet achieves competitive performance across various visual tasks, offering new insights for the development of physics inspired model architecture design.

\bibliographystyle{named}
\bibliography{ijcai25}

\end{document}